**Physics-Informed Machine Learning Regulated by Finite Element Analysis for Simulation Acceleration of Melt Pool Dynamics in Laser Powder Bed Fusion**

R. Sharma[a,b], Y.B. Guo[a,b*]

[a] Dept. of Mechanical and Aerospace Engineering, Rutgers University-New Brunswick, Piscataway, NJ 08854, USA
[b] New Jersey Advanced Manufacturing Institute, Rutgers University-New Brunswick, Piscataway, NJ 08854, USA

*Corresponding author: yuebin.guo@rutgers.edu (Y. Guo)

## Abstract

Efficient simulation of Laser Powder Bed Fusion (LPBF) is crucial for process prediction due to the lasting issue of high computational cost associated with traditional numerical methods such as finite element analysis (FEA). While a Physics-Informed Neural Network (PINN) can predict solution fields with small training data and enables the generalization of new process parameters via transfer learning, it suffers from accuracy degradation in time-dependent problems due to the accumulation of residual and the difficulty in capturing the steep spatial and temporal gradients inherent in the LPBF process. To overcome this issue, this study develops an efficient modeling framework*, FEA-Regulated Physics-Informed Neural Network (FEA-PINN),* to accelerate the prediction of melt pool dynamics phenomena in an LPBF process while maintaining the FEA accuracy. The innovation of *FEA-PINN* manifested itself in two aspects. First, a novel strategy has been developed within the PINN model to capture the dynamic phase change of powder-liquid-solid, enabling the tracking of material status during laser melting. The model further incorporates temperature-dependent material properties, phase change behavior of the powder bed, Marangoni convection, and natural convection within the melt pool. Second, the *FEA-PINN* framework integrates corrective FEA simulations during inference to enforce physical consistency, reduce error drift, and capture the steep gradients. A comparative analysis shows that *FEA-PINN* achieves accuracy comparable to FEA while significantly reducing computational cost. The framework has been validated against benchmark FEA data for single-track scanning in LPBF.

## 1. Introduction

Metal AM may produce complex components across various industries, including aerospace, biomedical, and tooling [1]. As a primary metal AM process, laser powder bed fusion (LPBF) has the potential to produce components with a near-net shape. A moving laser layer-by-layer scans a layer of 30-50 μm over a solid substrate in an inert gas environment. Despite the potential of LPBF, it has not been fully adopted in the industry due to the complex process-microstructure-property relationship. The rapid heating-melting-cooling thermal dynamics in LPBF often result in printing defects such as porosity, lack of fusion, microcracks, and distortion [2]. Therefore, it becomes essential to simulate an LPBF process before actual component printing. The prediction of melt pool dynamics (e.g., temperature and velocity fields) in an LPBF process is essential as it significantly affects the process-microstructure-property relationship.

Many researchers have developed numerical models at different length and time scales to capture the process physics in LPBF [3–5]. These models were validated to different degrees using experimental data captured from in-situ and ex-situ monitoring methods. These models employed different numerical methods like finite difference, finite volume, and finite element methods to solve the governing partial differential equations (PDEs) of the physical phenomena on the spatial discretization and time steps. Li et al. [6] developed a 2D LPBF model based on the volume of fluid (VOF) method by accounting for the effects of surface tension on tracking the melt pool interface. However, the Marangoni and recoil forces were excluded. Gurtler et al. [7] presented a transient 3D simulation of LPBF using OpenFOAM. Although the model did not consider surface tension, Marangoni effects, or recoil forces, it captured the essential behaviors of laser-based melting processes. Tseng et al. [8] examined how surface tension, Marangoni convection, and recoil forces influence the melt pool shape in LPBF. They showed that Marangoni flow widens the melt pool because it helps transfer heat from the laser's center outward, while surface tension causes the melt pool surface to oscillate periodically. Baere et al. [9] created a numerical model combining macro-scale thermal and fluid dynamics with a metallurgical model. This integrated approach predicted solidification parameters like temperature gradients, cooling rates, and grain growth speeds, determining whether the grains formed were columnar or equiaxed. The predictions were then used as the input in a microstructural model to simulate the final microstructure after heat treatment. In summary, these numerical models face three main challenges. First, they rely on simplified assumptions that differ from real-world process physics, causing inaccuracies. Second, model parameters need to be continuously updated due to the rapidly changing process dynamics during an LPBF process. Third, these simulations are costly and time-consuming, especially when dealing with multiple scales and physics interactions.

Nowadays, with more sensor data available during manufacturing, data-driven methods like machine learning (ML) (especially deep learning (DL)) offer a promising alternative way to uncover hidden patterns and accurately predict thermal behaviors in an LPBF process. These data-driven models predict the physical phenomena with comparable accuracy to the physics-based simulation models. ML models use algorithms to learn from data and predict outcomes, aiming for accuracy in tasks like classification and regression. The key advantage is flexibility, which allows for adapting to new instances easily once trained on one instance. This adaptability makes them ideal for real-time process predictions, unlike traditional physics-based simulations that are always computation-intensive when even a single process parameter changes. Researchers have attempted to predict the thermal history of the metal AM process using DL methods [10–13]. Although data-driven models are promising and effective for solving complex Multiphysics problems, their accuracy depends heavily on sufficient labeled training data. Unfortunately, obtaining enough training data through experiments or simulations can be extremely costly, especially for manufacturing processes such as metal AM.

Recently, the problem of a lack of labeled training data has been addressed by incorporating the governing PDEs into ML models. This branch of scientific ML is called Physics-informed neural network (PINN) [14]. Over the years, PINN methodology has undergone significant improvements, establishing itself as a promising alternative to conventional physics-based simulation models for the following reasons:

- The PINN method is a mesh-free approach, avoiding the high computational costs typically associated with numerical methods like the finite volume method (FVM), finite difference method (FDM), etc.
- PINN uses automatic differentiation, providing accurate derivatives without truncation errors inherent in traditional numerical techniques.
- PINN allows transfer learning, meaning once trained with one instance, it can rapidly predict outcomes for a family of new instances with minimal training cost, unlike numerical methods that require significant time to solve even for a single parameter change.

PINNs offer several advantages and have been successfully applied to various computational problems, including those involving nonlinear conservation laws such as fluid dynamics, biology, and heat transfer phenomena [15–17]. However, the original PINN approach has limitations, especially for complex problems. Several advanced versions have been introduced to address these issues and improve computational efficiency and accuracy. For instance, conservative PINNs (cPINNs) [17] divide the computational domain into multiple sub-domains, each represented by its own neural network (NN), enhancing accuracy in solving conservation laws. Similarly, the extended PINN (XPINN) [18] utilizes a generalized domain decomposition technique that allows the partitioning of the computational domain flexibly, leading to better model accuracy and computational performance. Furthermore, adaptive PINN strategies like transfer learning have significantly reduced computational costs by reusing previously trained models for predicting new instances.

In the last few years, researchers have developed PINN models to predict the thermal history of metal AM [19–22]. Zhu et al. [23] utilized the PINN model to predict thermal behavior and melt pool dynamics in the directed energy deposition (DED) process. They demonstrated that the model effectively captured these phenomena with fewer training data points than traditional methods. Additionally, their study highlighted that PINNs could achieve better predictions by implementing boundary conditions in a "hard" manner rather than a "soft" way. Liao et al. [24] also predicted the thermal history in the DED process using a PINN model. After training the model, they used experimental data on the top surface of the melt pool to predict the thermal field across the entire domain. They also predicted difficult-to-obtain material properties through inverse learning. Sajadi et al. [25] introduced a Physics-Informed Convolution LSTM (PI-ConvLSTM) framework for predicting 2D temperature fields in DED, notable for integrating in-situ temperature data, a customized physics-informed loss, and a physics-informed input. This framework demonstrates high accuracy and adaptability across various geometries, deposition patterns, and process parameters. Peng and Panesar [26] presented a PINN-based framework for multi-layer thermal simulation in DED, explicitly addressing the challenge of discontinuities between layers using strategies like pointwise weight assignment for initial conditions and explicit solid-void definitions. Similarly, other researchers have used PINNs to determine thermal history, melt pool dynamics, and thermal stress evolution, but exclusively for the DED process [27,28].

Despite these advances, the potential of PINNs for modeling the LPBF process remains largely unexplored due to three main challenges: (1) the length and time scales for LPBF processes are in the order of micrometers ($\mu m$) and microseconds ($\mu s$), respectively. Due to the many source terms involved in the physical phenomena, such as laser heat input, convective and radiative heat losses, and phase changes, it is difficult to normalize the governing equations. The small scale can lead to gradient vanishing problems during backpropagation when working with the original dimensions. (2) It is challenging to capture changes in material properties when powder melts into liquid and then solidifies to bulk using a fully connected NN, which could not capture time-dependent phenomena. (3) Error accumulation in long-term simulations may cause the PINN prediction trajectory to diverge from the ground truth [29,30]. Unlike first-principles solvers, NNs do not perform internal iterations to solve the governing equations and reduce residuals explicitly. As a result, the nonlinear PDEs with source terms and boundary conditions can only be accurately approximated within a limited time window close to the training data, where the combination of linear operations and activation functions remains effective. Despite significant advancements in PINN methodologies, they have not yet entirely replaced first-principles-based solvers. However, PINNs hold a strong potential to accelerate the performance of conventional physics-based solvers (FEA, FVM, etc.) when integrated strategically[31]. As such, there is a growing need for the development of a practical CFD acceleration framework that effectively meets the following three objectives: (1) accurate prediction of unseen (future) time series in long-term CFD simulations for complex problems like thermal field evolution and melt pool dynamics in LPBF, (2) computational performance that matches that of conventional CFD solvers, and (3) a reasonable requirement for training data and training time for a set of different process parameters.

To address the challenges associated with long-duration thermal and melt pool dynamics predictions in traditional FEA solvers, this study aims to develop an *FEA-regulated PINN (FEA-PINN)* framework for accurately predicting melt pool dynamics with significantly reduced computational cost. The proposed PINN model accounts for temperature-dependent material properties, phase change behavior of the powder bed, Marangoni convection, and natural convection in the melt pool. Upon solidification, the model updates the material state, assigning bulk solid properties to regions previously occupied by molten powder. The predictions obtained from the PINN model are compared against high-fidelity FEA simulations to evaluate accuracy. To address the error accumulation associated with long-duration PINN, the PINN model is integrated with an FEA solver, forming a hybrid FEA-PINN framework. This paper is organized as follows: *Section 2* illustrates the working principle of the FEA-PINN framework through a representative example and demonstrates how the integration of corrective FEA with PINN inference can enhance solution accuracy and computational efficiency in long-term simulations. *Section 3* introduces the governing equations relevant to the LPBF process. *Section 4* presents the architecture of the PINN model and describes the algorithm used to handle the time-dependent transition of material properties before and after melting. *Section 5* demonstrates the ability of the PINN to predict the temperature field and melt pool dynamics using limited training data and evaluates its capabilities to predict long-duration solutions. Then, finally, the FEA-PINN framework's role in improving long-term prediction accuracy through hybridization with FEA is highlighted.

## 2. FEA-regulated physics-informed neural network (FEA-PINN) strategy

In this study, the proposed *FEA-PINN* framework introduces a hybrid simulation approach that combines the efficiency of PINNs with the physical reliability of FEA to accelerate the simulation of long-duration and time-dependent complex LPBF processes. Fig. 1 illustrates a step-by-step implementation of this strategy using a representative example for one such problem. For instance, if a time-dependent simulation needs to be performed over a time span from 0 to 600 $\mu s$, the *FEA-PINN* framework applies to the sequence of training, prediction, correction, and acceleration in a staged manner across the time span to balance computational accuracy and efficiency.

The process begins with the generation of high-fidelity training data using an FEA solver for a short time interval, e.g., t = 0 to 120 $\mu s$. This data then trains the initial PINN model, which learns the physical phenomena governed by the PDEs and associated boundary conditions. Once trained, the PINN is deployed in an inference mode to predict the solution for future time steps. In this representative example, the model demonstrates reasonably accurate predictive capability over the interval of t=120 to 280 $\mu s$ given the supplied training data. However, as shown by red blocks in Fig. 1, the accuracy of PINN prediction begins to degrade between t=280 and 300 $\mu s$, due to residual accumulation when extrapolating far beyond the original training range. This initiates the correction phase, during which the PINN-predicted temperature field at t=280 $\mu s$ is used as the initial condition for a short FEA simulation that runs from t = 280 to 300 $\mu s$. During this interval, the FEA solver, using its internal iterative scheme, reduces the residuals of the governing PDEs and enforces boundary conditions, thereby correcting accumulated errors and restoring physical consistency to the solution. It is worth noting that the precise instant (280 $\mu s$) at which the correction loop is initiated is not critical, since the FEA solver employs the predicted solution as a warm start only for the iterative solver used to solve the mass–stiffness matrix. As a result, an approximate choice of the time instant to start the correction loop is sufficient to ensure convergence. The only implication of delaying the initiation of the correction loop is that the solver may require a longer time to reach an accurate solution. For example, if the correction loop is initiated at 380 $\mu s$, the solver will take more than 20 $\mu s$ to achieve accurate predictions. This work triggers the correction loop when the maximum absolute temperature error within the domain reaches 500 K.

Then, the corrected temperature field obtained at t=300 $\mu s$ can be used to retrain the PINN model for a short period. Because PINN has already captured physical behavior previously, this update requires only a small number of training epochs, in contrast to the much longer initial training phase. The updated PINN is then used to predict the temperature field for the next extended interval, e.g., t=300 to 440 $\mu s$, which is marked as the acceleration zone (yellow region in Fig. 1). In this zone, the model benefits from the speed of data-driven inference while maintaining physics-informed accuracy. This iterative loop may continue until the final time step.

In contrast to full-resolution FEA simulations across the entire time span, the *FEA-PINN* framework significantly reduces computational cost by limiting expensive FEA evaluations to brief correction intervals only when necessary. The duration and frequency of FEA correction steps may vary depending on the extent of error in the PINN's extrapolated predictions, the complexity of the physics involved (e.g., phase change, nonlinear properties), and the convergence behavior of the chosen FEA solver. Overall, the *FEA-PINN* framework provides a powerful tool for long-duration simulations in computationally expensive problems like thermal field prediction in LPBF. Its modular nature, combining offline FEA solver with online PINN inference, makes it well-suited for extension to multi-physics simulations and real-time digital twins of manufacturing processes. The working process of the *FEA-PINN* framework is demonstrated in the following section through the prediction of thermal field and melt pool dynamics in the LPBF process during a single-track scan. The results highlight how the *FEA-PINN* strategy improves solution accuracy and efficiency compared to a standalone PINN model.

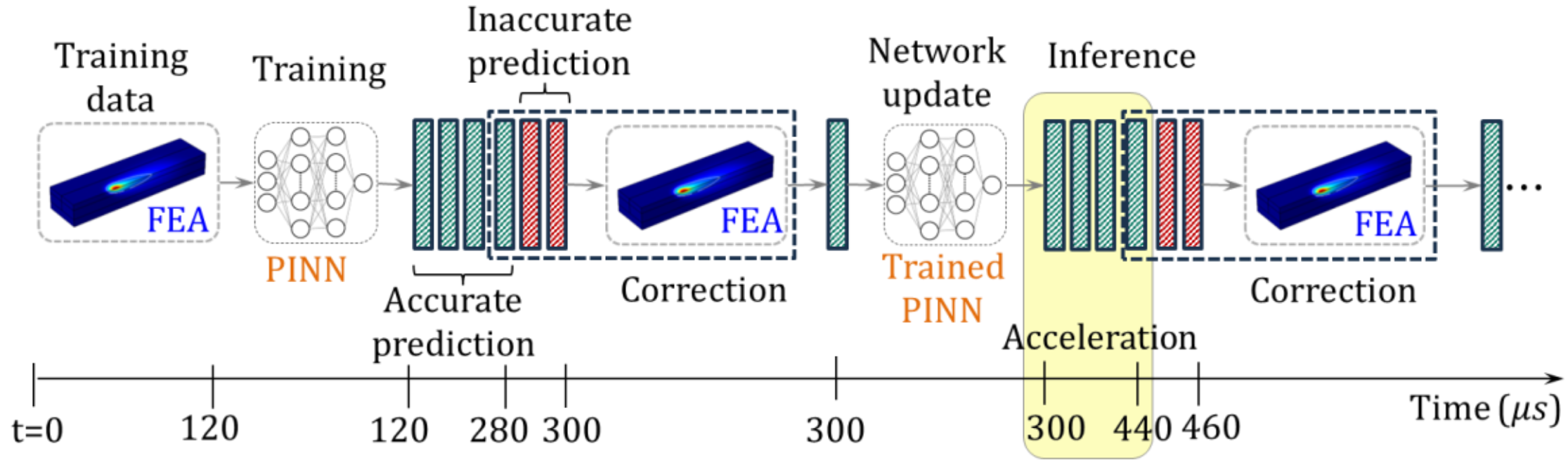

**Fig. 1:** Step-by-step implementation of FEA-PINN strategy (with an AM case).

## 3. Numerical model

This work focuses on the evolution of the thermal field and melt pool dynamics during a single-track scan of SS 316L powder over a substrate of the same material in the LPBF process. Fig. 2 shows a schematic representation of this process. When the laser begins scanning, the powder undergoes a phase change. A liquid melt pool forms as the laser heat builds up. Once the laser moves forward, the melt pool starts solidification by dissipating heat to the surroundings through convection and radiation and attains the material properties of the solid substrate (Fig. 2b).

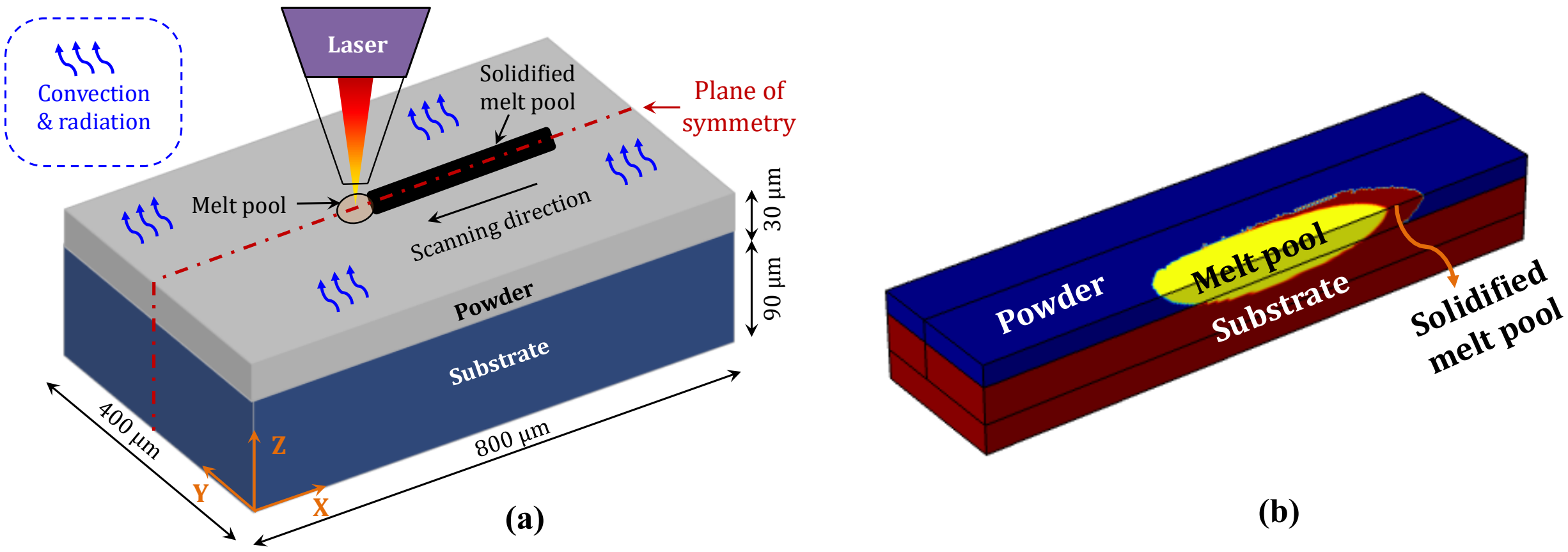

**Fig. 2:** Schematic of **(a)** LPBF **(b)** four different material phases co-existing in the simulation domain.

Certain assumptions were introduced to reduce complexity while still accurately capturing essential physics:

- The powder layer is considered a homogeneous layer with uniform material properties.
- Evaporation effects are not considered in this model.

The gas-filled pores and powder size distribution in the powder layer control material properties like thermal conductivity and density [32]. Considering this fact, the thermal conductivity and density of the powder layer can be calculated from the bulk solid and can be given as:

$$\rho_P = (1-\varphi)\rho_S \tag{1}$$

$$k_P = k_S \frac{(1-\varphi)}{(1+11\varphi^2)} \tag{2}$$

where, $\rho_P$ is the powder layer density, $\rho_S$ is the bulk solid density, $k_P$ is the powder layer thermal conductivity, $k_S$ is the bulk solid thermal conductivity, and $\varphi$ is the porosity and assumed to be 35% in this study. It is worth noting that while some of the governing equations used in this study are simplified (e.g., evaporation is not considered), the primary objective is to demonstrate the potential of the proposed *FEA-PINN* strategy to accelerate thermal field and melt pool dynamics simulations in the LPBF process. This study serves as a foundational benchmark, illustrating the effectiveness of the hybrid framework. Future research can build upon this work by incorporating additional physical complexities, such as keyhole dynamics, recoil pressure, and multi-track/multi-layer interactions, to enhance the model's fidelity and broaden its applicability.

### *3.1 Governing equations*

The governing equation consists of mass, energy, and momentum transport equations with suitable source terms and boundary conditions solved in the bulk solid and powder layer. This can be defined as:

$$\frac{\partial \rho}{\partial t} + \nabla \cdot (\rho \vec{u}) = 0 \tag{3}$$

$$\frac{\partial(\rho C_p T)}{\partial t} + \vec{u} \cdot (\rho C_p T) = \nabla \cdot (k \nabla T) \tag{4}$$

where, $\vec{u}$ is the molten metal flow velocity, $T$ is the temperature in the domain, $\rho$, $C_p$, and $k$ are the effective density, heat capacity, and thermal conductivity, respectively. These material properties are a function of the region (powder layer or bulk solid) and the material phase (powder, liquid, or solid). In the melting zone, the value of effective $\rho$, $k$, and $C_p$ are obtained by averaging the solid and liquid phases as follows:

$$\rho = (1-f_L)\rho_S + f_L \rho_L \tag{5}$$

$$k = (1-f_L)k_S + f_L k_L \tag{6}$$

$$C_p = \frac{1}{\rho}\{(1-f_L)\rho_S C_{pS} + f_L \rho_L C_{pL}\} + L\frac{\partial \alpha_m}{\partial T} \tag{7}$$

Here, the subscripts *'S'* and *'L'* correspond to solid and liquid phases. $L$ is the latent heat of phase change, and its effect is incorporated by the apparent heat capacity method in this study. $f_L$ and $\alpha_m$ are the liquid fraction and mass fraction, respectively, and can be given as:

$$f_L = \begin{cases} 0 & T < T_S \\ \dfrac{T - T_S}{T_L - T_S} & T_S \le T < T_L \\ 1 & T \ge T_L \end{cases} \tag{8}$$

$$\alpha_m = \frac{1}{2}\frac{(f_L \rho_L - (1-f_L)\rho_S)}{((1-f_L)\rho_S + f_L \rho_L)} \tag{9}$$

where, $T_S$ and $T_L$ are the solidus and liquidus temperature of the SS 316L.

The thermal boundary conditions are given by:

$$-k\,\frac{\partial T}{\partial n} = Q_{laser} + Q_{conv} + Q_{rad} \tag{10}$$

where $n$ is the normal to the surface, $Q_{laser}$ is the heat input by the laser heat source, $Q_{conv}$ is the convective heat loss and $Q_{rad}$ is the radiative heat loss and given by:

$$Q_{laser} = -\frac{2\eta P}{\pi r_b^2}\exp\left(\frac{-2\left((x-vt)^2+y^2\right)}{r_b^2}\right) \tag{11}$$

$$Q_{conv} = h\,(T - T_0) \tag{12}$$

$$Q_{rad} = \sigma\epsilon\,(T^4 - {T_0}^4) \tag{13}$$

where $\eta$ is the laser absorption coefficient, $P$ is the laser power, $r_b$ is the laser beam radius, $v$ is the laser scanning velocity, $h$ is the convective heat transfer coefficient, $\sigma$ is the Stefan-Boltzmann constant, $\epsilon$ is the emissivity and $T_0$ is the ambient temperature. The bottom surface of the substrate has a finite temperature boundary condition. The initial temperature of the domain is equal to the ambient (293.15 K).

The momentum conservation equation is given as:

$$\frac{\partial(\rho\vec{u})}{\partial t} + \vec{u}\cdot\nabla(\rho\vec{u}) = -\nabla p + \nabla\cdot\mu(\nabla\vec{u} + (\nabla\vec{u})^T) + \vec{F}_d + \vec{F}_b \tag{14}$$

where, $p$ is the pressure and μ is the dynamic viscosity of the liquid metal. The terms $\vec{F}_d$ and $\vec{F}_b$ are source terms representing Darcy drag force, incorporating mushy region flow resistance during liquid-solid phase change, and buoyancy force, respectively. It is given as:

$$\vec{F}_d = -\frac{(1-f_L)^2}{{f_L}^3 + \omega}C\vec{u} \tag{15}$$

$$\vec{F}_b = \rho_L\vec{g}\beta_T(T - T_{ref}) \tag{16}$$

here, $C$ and $\omega$ are constants and are given the values of $1.6\times 10^6$ $kg.m^{-3}s^{-1}$ and $10^{-3}$ respectively[33]. $\vec{g}$ is the acceleration due to gravity and $\beta_T$ is the coefficient of thermal expansion, and $T_{ref}$ is the reference temperature (293.15 K).

The following equations describe the Marangoni flow induced on the top surface due to the steep temperature gradients generated during scanning with the Gaussian laser heat source. With $\sigma$ as the surface tension coefficient, this effect is incorporated into the model as a boundary condition:

$$\tau_{xz} = \mu\frac{\partial u}{\partial z} = \left(\frac{\partial\sigma}{\partial T}\right)\cdot\left(\frac{\partial T}{\partial x}\right) \tag{17}$$

$$\tau_{yz} = \mu\frac{\partial v}{\partial z} = \left(\frac{\partial\sigma}{\partial T}\right)\cdot\left(\frac{\partial T}{\partial y}\right) \tag{18}$$

The different process parameters and temperature-dependent material properties used in the study are tabulated as follows:

Table 1. Process parameters and material properties of SS 316L [34].

| Parameter | Value |
| --- | --- |
| Laser power, $P$ (W) | 100 |
| Laser absorption coefficient, $\eta$ | 0.4 |
| Laser beam radius, $r$ ($\mu$m) | 40 |
| Laser scanning speed, $v$ (mm/s) | 800 |
| Porosity of powder ($\varphi$) | 0.35 |

| Material property | Value |
| --- | --- |
| Solidus temperature, $T_S$ (K) | 1658 |
| Liquidus temperature, $T_L$ (K) | 1723 |
| Solid density, $\rho_S$ (kg/m$^3$) | $8084 - 0.4209\,T - 3.894 \times 10^{-5}\,T^2$ |
| Liquid density, $\rho_L$ (kg/m$^3$) | 6873 |
| Solid heat capacity, $C_{pS}$ (J/kgK) | $462 + 0.134\,T$ |
| Liquid heat capacity, $C_{pL}$ (J/kgK) | 775 |
| Solid thermal conductivity, $k_S$ (W/mK) | $9.248 + 0.01571\,T$ |
| Liquid thermal conductivity, $k_L$ (W/mK) | 22.5 |
| Latent heat, $L$ (J/g) | 270 |
| Convective coefficient, $h$ (W/m$^2$K) | 40 |
| Thermal expansion coefficient, $\beta_T$ (K$^{-1}$) | $1.1\times 10^{-5}$ |
| Surface tension gradient, $\frac{d\sigma}{dT}$ (Nm$^{-1}$K$^{-1}$) | $-0.4 \times 10^{-3}$ |
| Dynamic viscosity, μ (Nm$^{-1}$s$^{-1}$) | $6 \times 10^{-3}$ |
| Emissivity, $\epsilon$ | 0.26 |

*3.2 Physics-based simulation model*

To validate the results of the data-driven model, generate the training data, and improve the accuracy of the hybrid FEA-PINN model, COMSOL Multiphysics®, a commercial software, is used. The generated data is treated as the benchmark for this study, and a small portion is used to train the neural network. A substrate of dimensions 800 $\mu$m × 200 $\mu$m × 90 $\mu$m with 30 $\mu$m thick powder layer over it is scanned by a 100 W laser heat source with a scanning speed of 800 mm/s (Fig. 2a). Due to geometric and thermal symmetry in single track scanning, only half of the computational domain is considered in the simulation to reduce computational cost without compromising accuracy. As the laser irradiates the powder, a phase change occurs, transforming the powder into a liquid. When the laser moves forward from that initial position, the melt pool solidifies by losing heat to the environment through convection and radiation. When this occurs, the liquid melt pool solidifies and attains the material properties of the bulk solid, as shown in Fig. 2b. A hexahedral mesh element was used to discretize the computation domain. To improve computational efficiency, mesh refinement was applied in the region where the laser interacts with the powder bed (because the laser path is already known) and high temperature gradients are present (Fig. 3a). In the remaining regions of the domain, where the temperature gradient is negligible, a coarser mesh is applied to reduce computational cost without compromising accuracy. A grid independence test was performed to eliminate the variation in results due to mesh size. Based on the grid independence test, 13620 elements were used to discretize the domain, which gives the minimum element size of 7.6×10×12 $\mu$m. The simulation was done for 600 $\mu$s. The simulation took approximately 11.75 hours on an Intel Core i7 CPU.

Similar to the mesh refinement in physics-based simulations, the selection and distribution of collocation points are critical for accurately capturing solution behavior in PINN modeling. A high density of collocation points was allocated near the top corner of the domain with steep temperature gradients where the laser scans the material, as shown in Fig. 3b. A coarser distribution of collocation points was used in the remaining domain to optimize computational resources while maintaining solution accuracy. The details of the PINN model will be discussed in the subsequent sections.

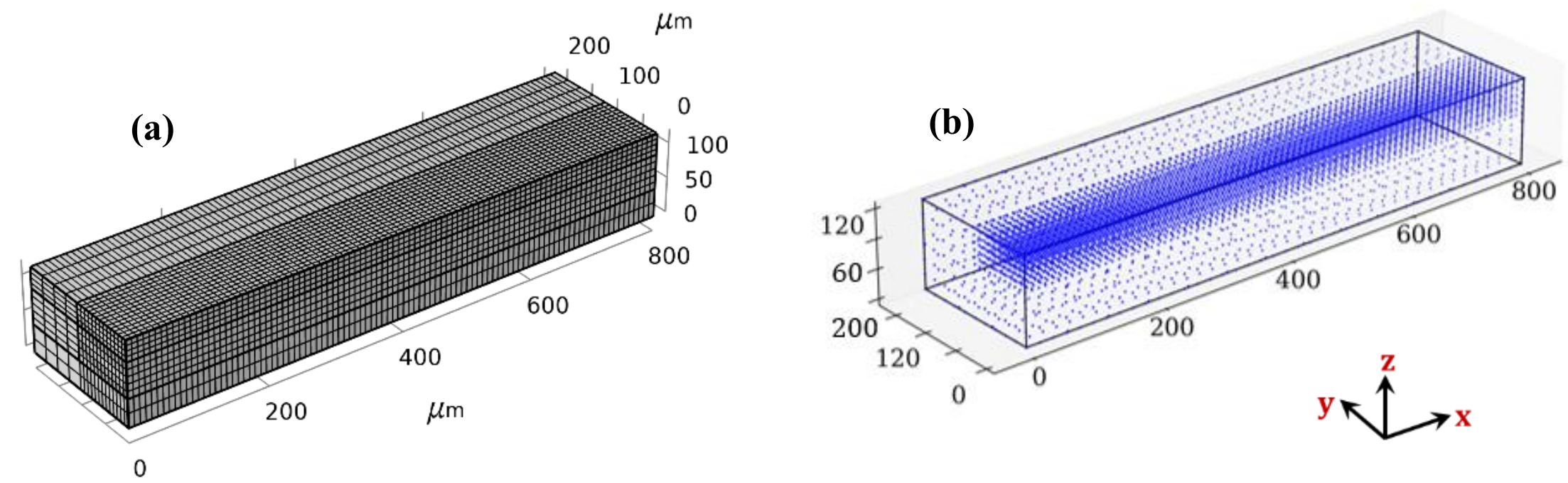

**Fig. 3:** (a) Meshing for FEA model (b) collocation points for PINN model

## 4. PINN Methodology

PINNs are a type of NN that utilizes the governing laws of physical phenomena as a constraint for training. The labeled training data is sometimes optional for PINNs (depends on the complexity of the problem), and due to this fact, PINNs are very effective because sometimes the generation of labeled data through experiments or simulation is quite expensive. The detailed working principle of a PINN model was explained in our previous study[35]. A PINN model uses automatic differentiation (AD) to compute the derivative, which is superior to the numerical derivative (using the Taylor series) as it does not contain truncation errors and fits very well on a GPU. In this study, the PINN model was implemented using PyTorch, and a relative $L_2$ error was calculated between the predicted and benchmark FEA data to measure the prediction accuracy.

### *4.1 PINN architecture*

In this study, two fully connected neural networks consisting of six hidden layers are employed to estimate the temperature and velocity fields across the entire computational domain. The architect of both networks is similar. The first and last hidden layers contain 32 neurons each, while the intermediate hidden layers comprise 64 neurons per layer. This architecture was finalized after extensive experimentation, beginning with networks ranging from 3 to 8 hidden layers. The number of neurons per layer varied between 16 and 64 to identify a configuration that balances training stability, accuracy, and efficiency. The input to the neural network is spatiotemporal coordinates $\{x, y, z, t\}$ and output is temperature for the temperature-network and velocity/pressure for the fluid-network (Fig. 4). The networks are randomly initialized using the Glorot method, and during the training of the network, the Adam optimizer [36] is used with a learning rate of $1e^{-3}$. The input of the network is scaled between -1 and 1. The tanh activation function is used for all the hidden layers except the last layer, where no activation function is used and kept as a linear layer. The Softplus activation function is used in the literature for the previous layer to predict the temperature. However, Softplus could not capture the phase change as it smoothens the sharp liquid-solid transition, which is incorrect. The PINN model considers extra loss terms apart from the data loss as given by the following equations:

$$\mathcal{L}_{data}^{T} = \frac{1}{M}[T_{Pred}(x, y, z, t) - T_{exact}(x, y, z, t)]^2 \tag{19}$$

$$\mathcal{L}_{data}^{F} = \frac{1}{M}[\vec{u}_{Pred}(x, y, z, t) - \vec{u}_{exact}(x, y, z, t)]^2 \tag{20}$$

$$\mathcal{L}_{PDE}^{T} = \frac{1}{N}\left[\frac{\partial(\rho C_p T)}{\partial t} + \vec{u} \cdot (\rho C_p T) - \nabla \cdot (k \nabla T)\right]^2 \tag{21}$$

$$\mathcal{L}_{PDE}^{F} = \frac{1}{N}\left[\frac{\partial(\rho \vec{u})}{\partial t} + \vec{u} \cdot \nabla(\rho \vec{u}) + \nabla p - \nabla \cdot \mu\left(\nabla \vec{u} + (\nabla \vec{u})^{T}\right) - \vec{F}_d - \vec{F}_b\right]^2 \tag{22}$$

$$\mathcal{L}_{BC}^{T} = \frac{1}{P}\left[Q_{laser} + Q_{conv} + Q_{rad} + k\frac{\partial T}{\partial n}\right]^2 \tag{23}$$

$$\mathcal{L}_{BC}^{F} = \frac{1}{P}\left[\mu\frac{\partial \vec{u}_t}{\partial z} - \frac{\partial \sigma}{\partial T}\nabla_t T\right]^2 \tag{24}$$

$$\mathcal{L}_{IC}^{T} = \frac{1}{Q}[T(x, y, z, 0) - T_0]^2 \tag{25}$$

$$\mathcal{L}_{IC}^{F} = \frac{1}{Q}[\vec{u}(x, y, z, 0) - \vec{u}_0]^2 \tag{26}$$

$$\mathcal{L}_{Total}^{T} = w_1.\mathcal{L}_{data}^{T} + w_2.\mathcal{L}_{PDE}^{T} + w_3.\mathcal{L}_{BC}^{T} + w_4.\mathcal{L}_{IC}^{T} \tag{27}$$

$$\mathcal{L}_{Total}^{F} = w_5.\mathcal{L}_{data}^{F} + w_6.\mathcal{L}_{PDE}^{F} + w_7.\mathcal{L}_{BC}^{F} + w_8.\mathcal{L}_{IC}^{F} \tag{28}$$

The superscript $T$ and $F$ refer to the temperature and fluid networks, respectively. $\mathcal{L}_{data}$ takes care of NN learning from the labeled training data, while $\mathcal{L}_{PDE}$, $\mathcal{L}_{Bc}$, and $\mathcal{L}_{IC}$ are the extra constraints in the PINN model that enforces the governing physical laws in the NN training. $M$, $N$, $P$, and $Q$ are the total number of sampling points for labeled data, governing PDE, boundary conditions, and initial conditions, respectively. The total loss comprises of the weighted sum of the four individual losses represented by respective $w$. The schematic of PINN architecture used in this study is represented in Fig. 4. The model was trained in ~95 minutes using an Nvidia RTX A6000 GPU and took 30000 epochs.

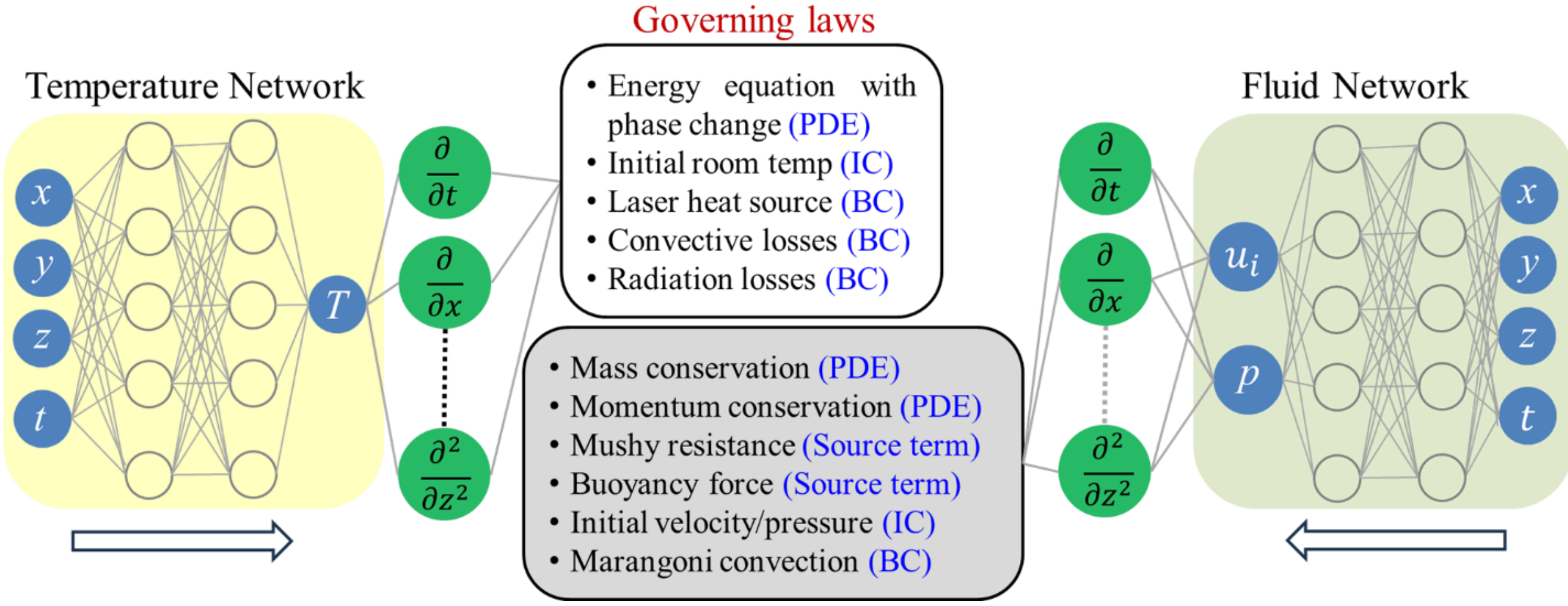

**Fig. 4:** Schematic of PINN model used in this study. Two fully connected neural networks are used to estimate the temperature and velocity field in the domain. The network for temperature has a yellow colour, while the network for fluid dynamics has a green colour. Automatic differentiation is used to compute the gradient.

### *4.2 Phase change and material properties update by PINN model*

In the literature, fully connected NNs are commonly employed in PINN frameworks [37]. However, accurately modeling powder-liquid-solid phase change and assigning temperature-dependent material properties, particularly within the powder layer of the domain, remains a significant challenge due to the co-existence of powder, liquid melt pool, and solidified melt pool (bulk solid) phases. Updating material properties that evolve over time is difficult to implement using a fully connected NN due to its limited ability to incorporate time-dependent phenomena.

As a result of this limitation, most PINN models developed to date focus on the DED process, where a powder layer does not exist in the simulation, and the phase interactions are less complex compared to LPBF. This study presents a strategy to incorporate the phase change and dynamic material property update within the PINN framework for LPBF. Algorithm 1 illustrates the pseudo-code to implement the strategy developed for this purpose. The yellow-highlighted section illustrates the core logic used for phase change tracking within the PINN framework. First, a binary state variable is defined for each spatiotemporal point based on the local temperature history. Points not exceeding the liquidus temperature ($T_L$) are assigned *State: 0*, while points exceeding $T_L$ at any time are assigned *State: 1* from the time step of first melting onward. This temporal tracking allows the model to differentiate between powder, liquid, and solid phases. Subsequently, material properties are updated dynamically based on both the current temperature and state value: powder for points below $T_S$ with *State: 0*, liquid for points above $T_L$ with *State: 1*, and bulk solid for points below $T_S$ with *State: 1*. For temperatures between the solidus ($T_S$) and liquidus ($T_L$) temperatures, the material properties are interpolated using Eq. (5–7), ensuring a smooth transition across the mushy zone.

**Algorithm 1: Pseudo-code for PINN of LPBF**

**Required:** The architecture of the neural network, hyperparameters, number of epochs, and material properties

1. Construct the networks with predefined parameters
2. Generate the sampling points in the domain and its boundaries
3. Initialize the neural networks
4. Forward propagation of PINNs to determine the temperature, velocities and pressure on sampling points
5. Calculate the derivative of temperature, velocities and pressure wrt *x,y,z,t* using automatic differentiation
6. Calculate the total loss given in Eqs. 27 and 28
7. Update the parameters of NNs based on Adam optimizer (Back-propagation)
8. Assign the "State"
   - If $T < T_L$ for a point *x,y,z,t* assign **"State:0"** to that point
   - If $T > T_L$ for a point *x,y,z,t* then find $t_{min}$ at which temperature exceeds the melting temperature
   - Once $t_{min}$ is determined, assign **"State:1"** to all the points [$x,y,z,(t > t_{min})$]
9. Assign material properties
   - If $T < T_S$ and "State:0" for a point *x,y,z,t* assign it as Powder
   - If $T > T_L$ and "State:1" for a point *x,y,z,t* assign it as Liquid
   - If $T < T_S$ and "State:1" for a point *x,y,z,t* assign it as Bulk solid
10. Repeat the training for the given number of epochs

**Output:** The temperature, velocities and pressure distribution in the domain

## 5. Results

### *5.1 Loss convergence*

Fig. 5 represents the evolution of training loss of the temperature and fluid PINN models for the training time duration. In this study, the weights associated with the individual loss components in Eq. (27) – namely $w_1$, $w_2$, $w_3$, and $w_4$ are set to $\{1, 1, 1, 1e^{-4}\}$ and Eq. (28) – namely $w_5$, $w_6$, $w_7$, and $w_8$ are set to $\{1, 1, 1, 1e^{-2}\}$. These weights are chosen in proportion to the magnitudes of their respective loss terms to ensure a balanced contribution during training. This strategy prevents any single term from dominating the loss function, thereby promoting stable and effective learning across all physical constraints.

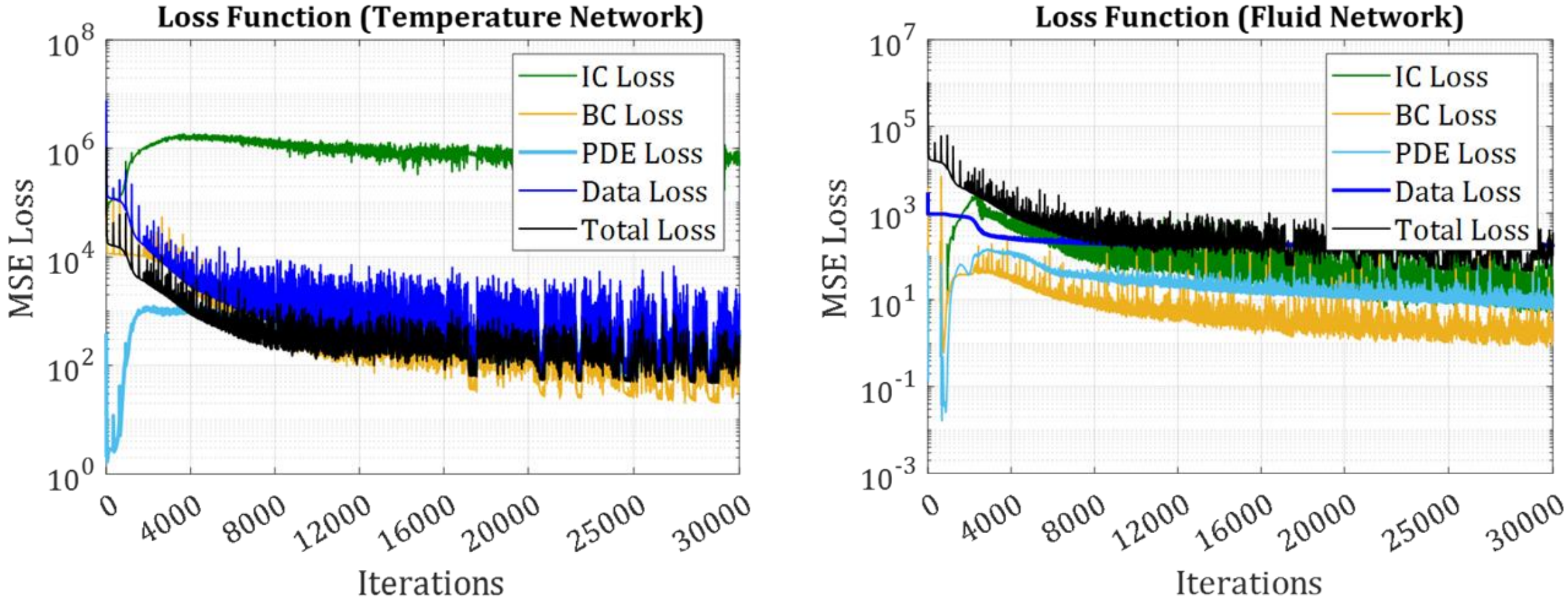

**Fig. 5:** Evolution of loss functions during model training **(a)** temperature network (**b)** fluid network

An interesting observation is the behavior of the PDE loss during the training of the temperature network. Initially, the PDE loss is very low, as the PINN model predicts a nearly uniform temperature field, which trivially satisfies the energy equation (Eq. 4). As training progresses and the model begins to capture spatial temperature gradients, the PDE loss increases due to a deviation from the true solution. Eventually, the PDE loss converges as the model learns the accurate temperature distribution that satisfies both the physical constraints and the data-driven targets. The same behavior is observed in the fluid network for PDE and boundary loss, where gradients are involved.

It is worth noting that PINN models in the literature have demonstrated the capability to learn physical behavior without relying on any training data. However, as the model complexity increases, particularly when incorporating physics such as phase change and the dynamic update of temperature-dependent material properties, buoyancy, and Marangoni convection on the top surface, a small set of training data becomes essential to guide convergence. In this study, the PINN model is provided with temperature and velocity data at ~21,000 spatial points per time step for three selected time instances: 40 $\mu s$, 80 $\mu s$, and 120 $\mu s$. This targeted inclusion of training data improves the stability and accuracy of the model, especially in capturing sharp gradients and non-linear phenomena associated with phase transitions.

*5.2 Material properties update*

Fig. 6 illustrates the spatial distribution of temperature-dependent material properties by the PINN model at t=600 $\mu s$ for various collocation points in the computational domain. The density plot (Fig. 6a) shows three distinct phases – powder, liquid, and solid – clearly identified within the top powder layer. Unscanned powder particles exhibit the lowest density, shown in blue, while the liquid melt pool is represented by the higher density (6873 kg/m$^3$) region in yellow. The solidified melt pool, which has undergone phase transition, is assigned the corresponding temperature-dependent bulk solid properties.

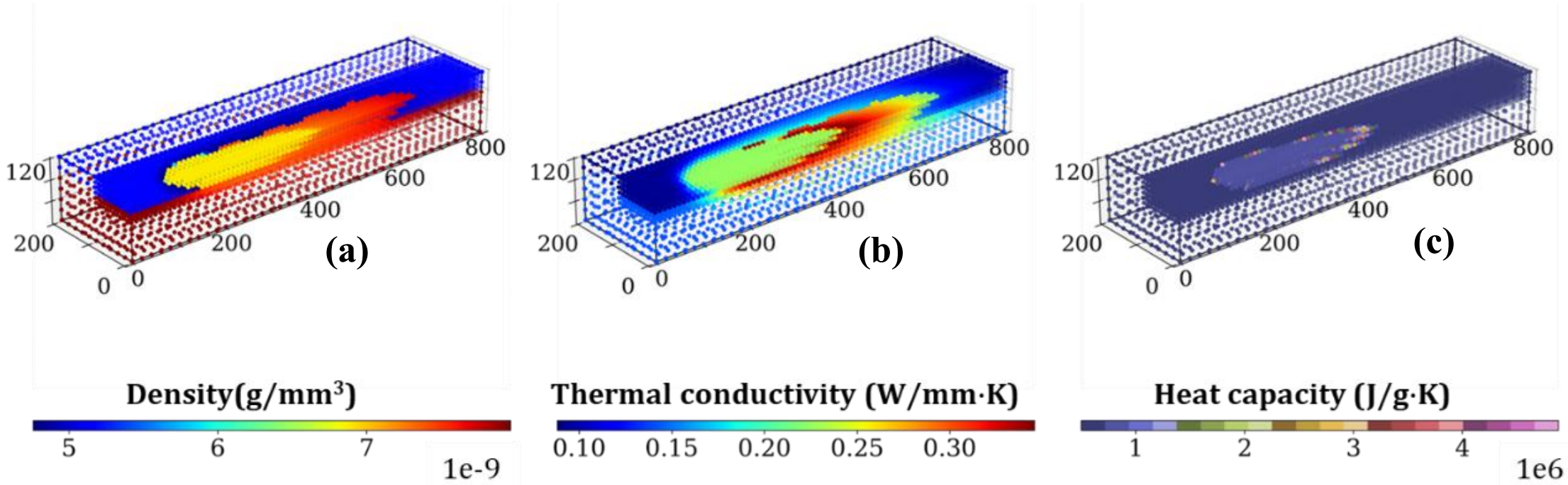

**Fig. 6:** **(a)** Density **(b)** thermal conductivity **(c)** heat capacity prediction by PINN model at t=600 $\mu s$

A similar phase-dependent behavior is observed in the thermal conductivity distribution (Fig. 6b), where values vary according to the local phase and temperature. The impact of latent heat during phase change is captured in the heat capacity plot (Fig. 6c). Specifically, the boundary of the melt pool exhibits significantly elevated heat capacity values, indicating the absorption of latent heat during the melting of powder. These spatially varying material properties are critical for accurately modeling heat transfer and phase evolution in the LPBF process.

*5.3 Temperature and velocity prediction*

Fig. 7 and Fig. 8 represent a comparison between the PINN predicted temperature and velocity fields and the benchmark FEA simulation at two distinct time instances: t=280 $\mu s$ and t=600 $\mu s$ respectively. The point-wise error distributions between FEA and PINN results at each time point are also shown. Despite being trained on sparse temperature data at three instances up to t=120 $\mu s$, the PINN model is capable of predicting the 3D temperature and velocity field in the LPBF process, including the steep thermal gradients near the laser heat source till t=600 $\mu s$. It is worth noting that, as the melt-pool dynamics are primarily driven by Marangoni convection on the top surface, the effects of the vertical velocity and pressure fields are negligible and are therefore not reported here.

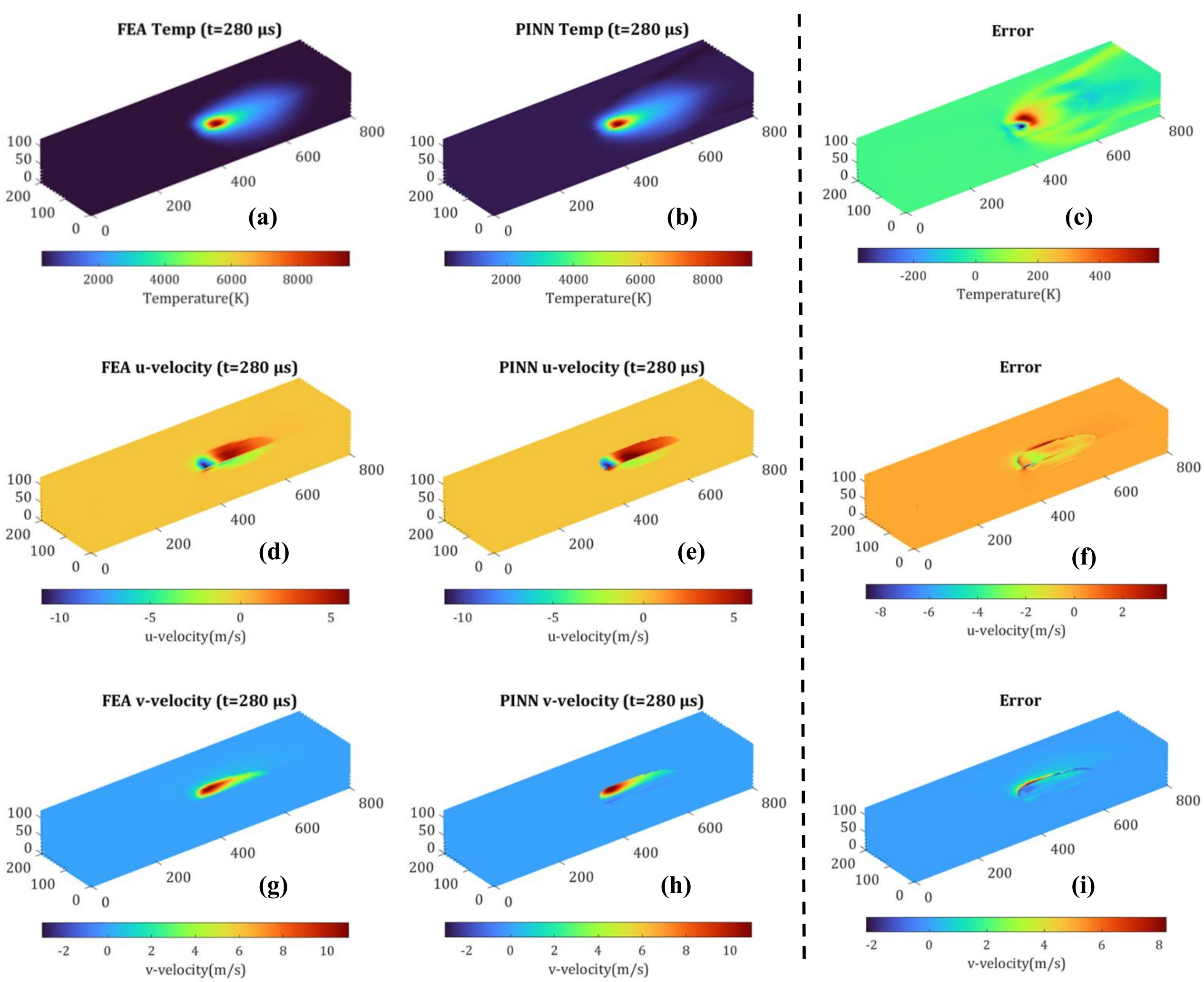

**Fig. 7:** Comparison of PINN-predicted temperature and velocity field with FEA results at t = 280 $\mu s$

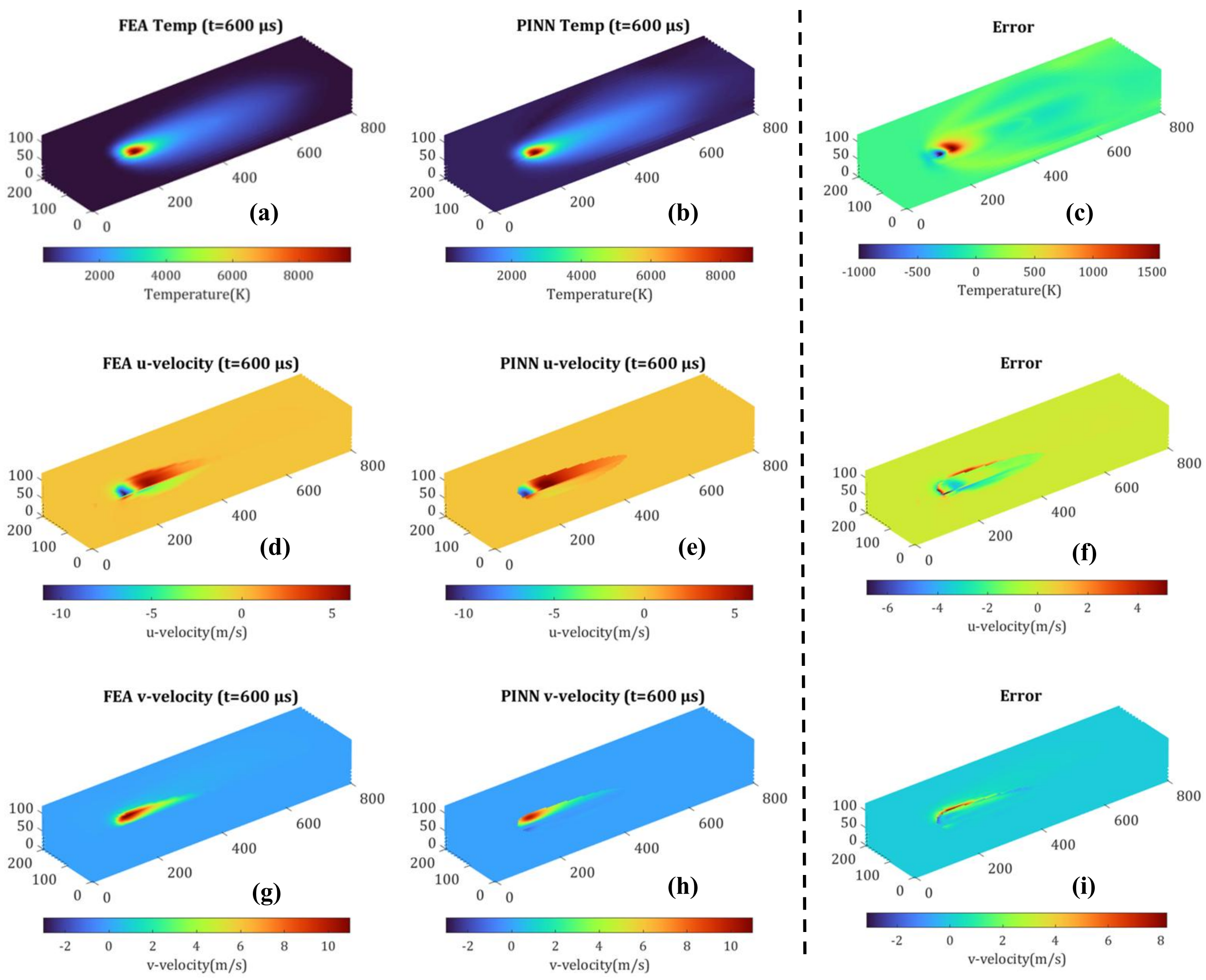

**Fig. 8:** Comparison of PINN-predicted temperature and velocity field with FEA results at t = 600 $\mu s$

At t=280 $\mu s$The predicted temperature field closely matches the FEA simulation, which is relatively close to the training data time-point. The point-wise error remains low across most domains with localized discrepancies at some points. This behavior is also illustrated in the line plot along the laser scan path at the top surface corner, where the PINN prediction captures both the peak temperature, thermal gradient, and fluid velocity with reasonable accuracy at t=280 $\mu s$ (Fig. 9a and 9b). However, the accuracy declines as the model extrapolates far beyond the training time range. At t=600 $\mu s$, a decrease in the max melt pool temperature and melt pool size is observed (Fig. 9c and 9d). It is observed that the PINN-predicted temperature field fails to fully resolve the steep thermal gradients at the melt pool boundary. Instead, the network tends to smooth out these sharp transitions, resulting in a slightly diffused thermal contour near the solid–liquid interface, as illustrated in Fig. 9(c). A significant error is observed in the u-velocity prediction (Fig. 9d).

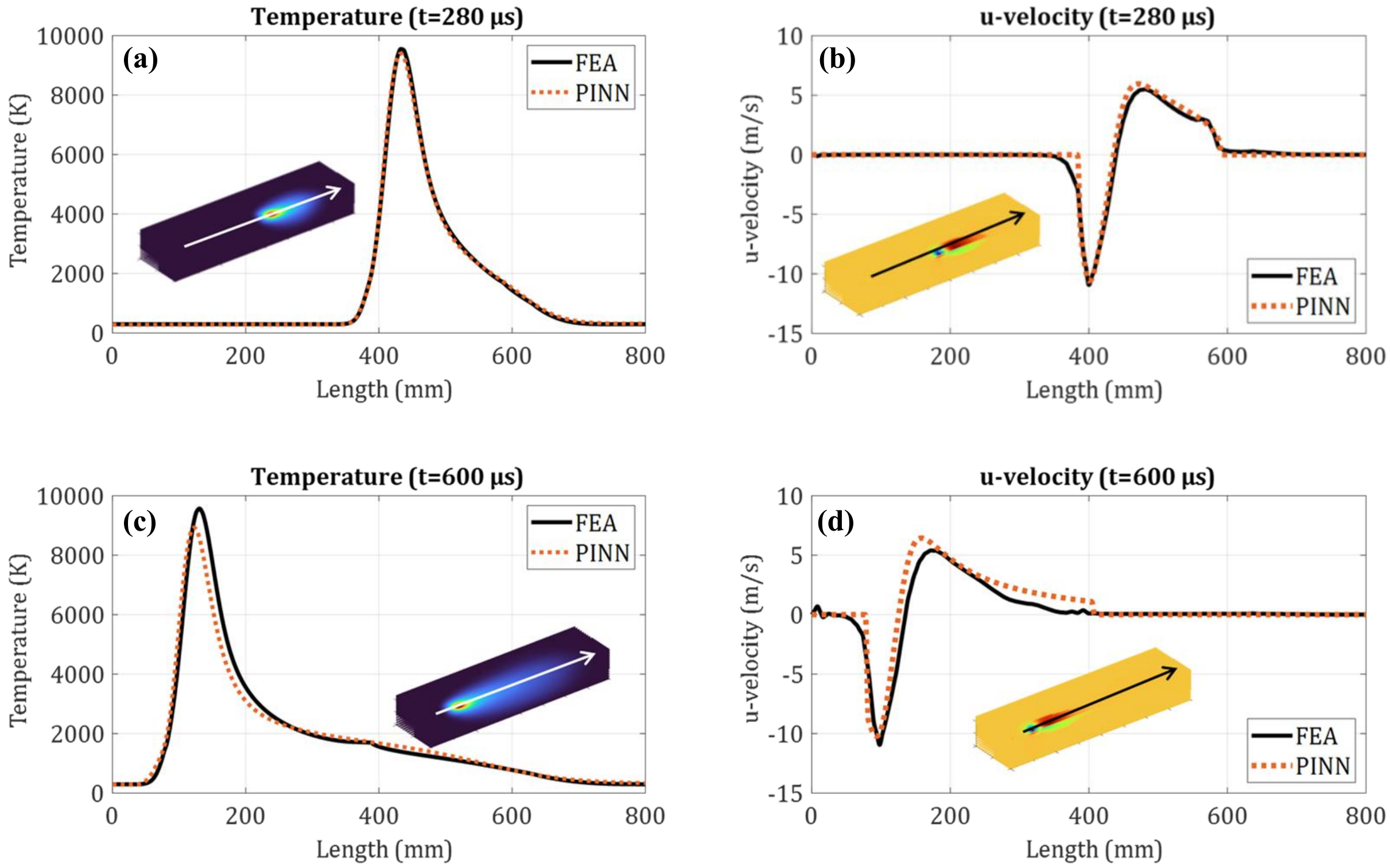

**Fig. 9:** Temperature distribution comparison along the laser scanning path at different time steps

This degradation in prediction accuracy over time highlights a limitation of PINN models when applied to time-dependent problems. Due to the inherently high temperature gradients involved, LPBF processes exhibit extreme thermal sensitivity, necessitating precise prediction of the temperature and velocity fields to capture the underlying melt pool dynamics and associated phase transformations with fidelity. Since PINNs learn in a continuous function space, their extrapolation ability diminishes as they move away from the supervised data points. This issue has also been observed in prior studies [38]. To address this limitation in long-term prediction accuracy, a unique hybrid approach combining the strengths of both PINN and FEA models is proposed in this study. This method leverages the data efficiency of PINNs and the temporal accuracy of FEA and is discussed in detail in the subsequent section.

*5.4 Parametric study with transfer learning*

One of the key advantages of data-driven ML models over traditional numerical approaches lies in their ability to generalize across different input conditions. Once trained on a specific set of process parameters, an ML model can be efficiently adapted to predict outcomes for a new set of parameters with minimal additional computational effort. This property makes such models particularly well-suited for soft sensing and real-time monitoring applications.

Fig. 10 illustrates the evolution of the loss function and mean square validation error comparing the PINN prediction with benchmark FEA data, with different combinations of laser power and scanning speed: 60 W at 600 mm/s. Among several test cases evaluated, only this case is presented here for clarity. The remaining parameter combinations exhibited similar trends and are therefore omitted for brevity. A benchmark PINN model, initially trained on a base case of 100 W at 800 mm/s, was employed as a pre-trained model for the new case. The transfer learning approach enables rapid convergence of the PINN model, with the temperature and melt pool velocity predictions stabilizing after approximately 3000 iterations in ~11 minutes, significantly faster than the time required for full model training (~95 minutes) using the same GPU resource.

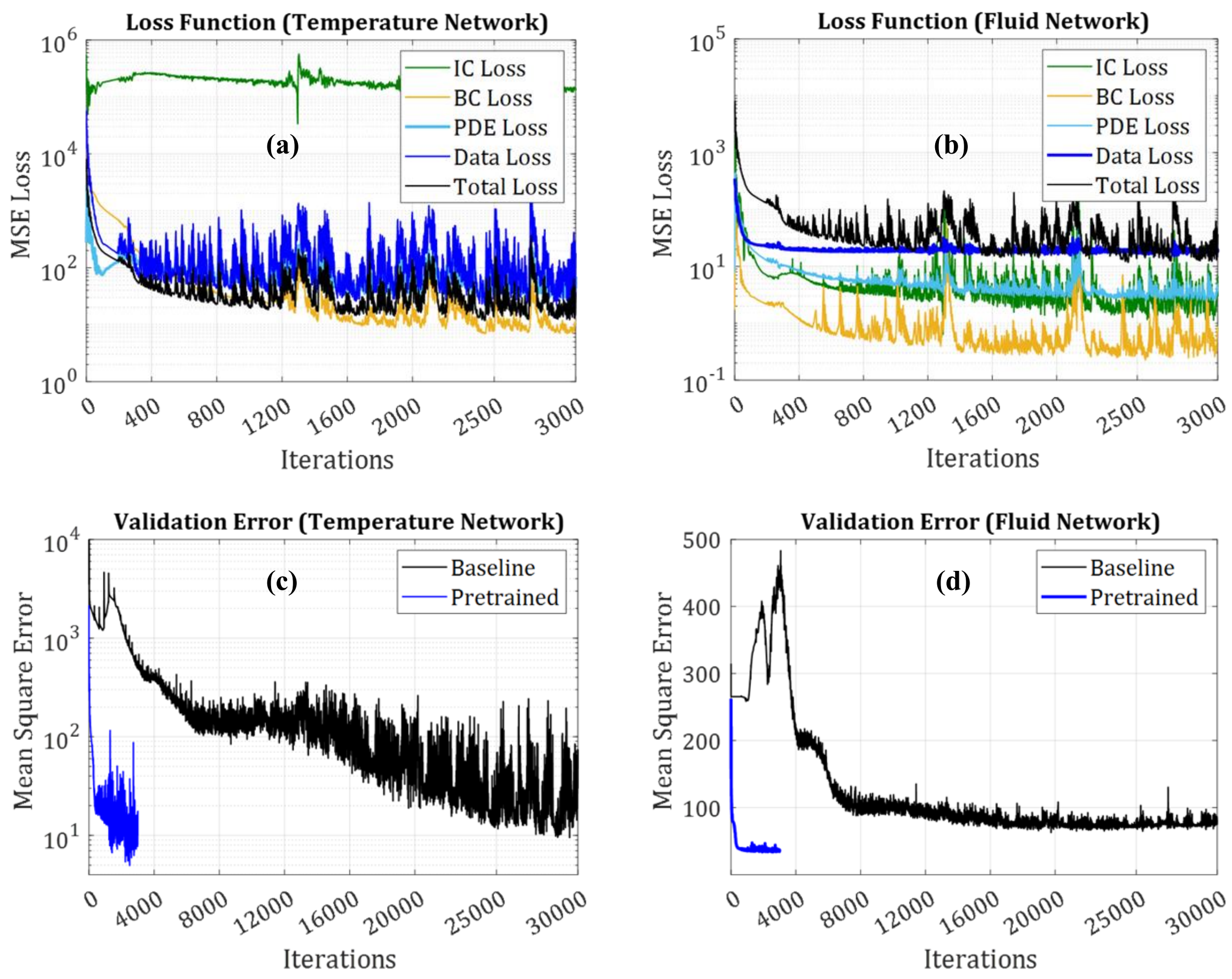

**Fig. 10:** Loss function and validation error for temperature and fluid network.

## *5.5 FEA-regulated PINN (FEA-PINN) model*

Previous studies in the literature have shown that a fully accurate solution, by eliminating the growth of prediction error over time, is unrealistic when using a single training phase for a data-driven model in time-dependent problems [29,30,38]. In conventional numerical simulations, internal iteration schemes are employed at each time step to ensure that the residuals of conservation equations remain within acceptable limits. In contrast, data-driven method-based time series predictions lack such iterative correction mechanisms for each time step, leading to the accumulation of residuals as time progresses. In physics-based models, a sudden increase in residuals is often an early indicator of convergence issues. When the solution begins to deviate from the governing equations, the error relative to the ground truth escalates rapidly. In data-driven models without iterative correction, the inability to suppress residual growth results in deteriorating accuracy over an extended time. Therefore, capturing long-term time series behavior using a single training process remains a significant challenge due to the cumulative nature of residual errors.

To address this challenge, an *FEA-regulated PINN (FEA-PINN)* framework is developed to accurately predict thermal field evolution with significantly reduced computational cost compared to traditional FEA solvers. The core idea behind the proposed hybrid strategy is to mitigate the error accumulation in PINN-predicted time series data by periodically invoking intermediate FEA simulations. As the PINN model extrapolates beyond the range of its training data, residuals in the predicted solution gradually increase. Once the discrepancy between the PINN prediction and the corresponding FEA solution exceeds a tolerable threshold, the PINN predicted field is used as the initial condition in the FEA solver, like COMSOL Multiphysics®. The FEA solver then computes the solution over a short future time window. Through its internal iterative scheme, the FEA solver effectively reduces the residuals of the governing heat equation and enforces boundary conditions, thereby correcting the accumulated errors in the PINN-predicted region and restoring physical consistency.

To demonstrate the correction capability of the proposed *FEA-PINN* strategy, the temperature, velocity, and pressure field predicted by the PINN model at t=280 $\mu s$ was used as the initial condition in a COMSOL-based FEA simulation. The model ran up to t=300 $\mu s$, taking approximately 15 minutes for this corrective window. During this interval, the FEA solver reduced the residuals of the heat equation and enforced boundary conditions through internal iterations, thereby correcting the accumulated errors and restoring physical consistency. It should be noted that the exact timing (280 $\mu s$) of the correction loop initiation is not critical, as the FEA solver uses the predicted field merely as a warm start for the iterative mass–stiffness solution. A later initiation only increases the convergence time. For instance, triggering the loop at 380 $\mu s$ instead of 280 $\mu s$, results in an additional delay of over 20 $\mu s$ before achieving accurate predictions. In this study, the correction loop is activated when the maximum temperature error in the domain reaches 500 K.

This correction and retraining process was repeated three times at t=280 $\mu s$, 440 $\mu s$, and 580 $\mu s$ throughout the prediction timeline. Each time, the corrected field from the FEA simulation was used to re-train the PINN model through transfer learning using a small number of epochs (3000 epochs). The updated PINN then resumed inference for the next ~160 $\mu s$ until the next correction was needed. This iterative hybrid loop successfully maintained high prediction accuracy while significantly reducing total simulation time. It is important to note that the duration required by the FEA solver to perform the correction depends on three factors: (1) the magnitude of the error in the PINN prediction, (2) the complexity of the underlying physics, and (3) the characteristics and convergence threshold of the FEA solver employed. In this particular case, a correction duration of 20 $\mu s$ (from 280-300 $\mu s$, 440-460 $\mu s$, and 580-600 $\mu s$) was sufficient to reduce the residuals to acceptable levels and realign the solution with the reference ground truth.

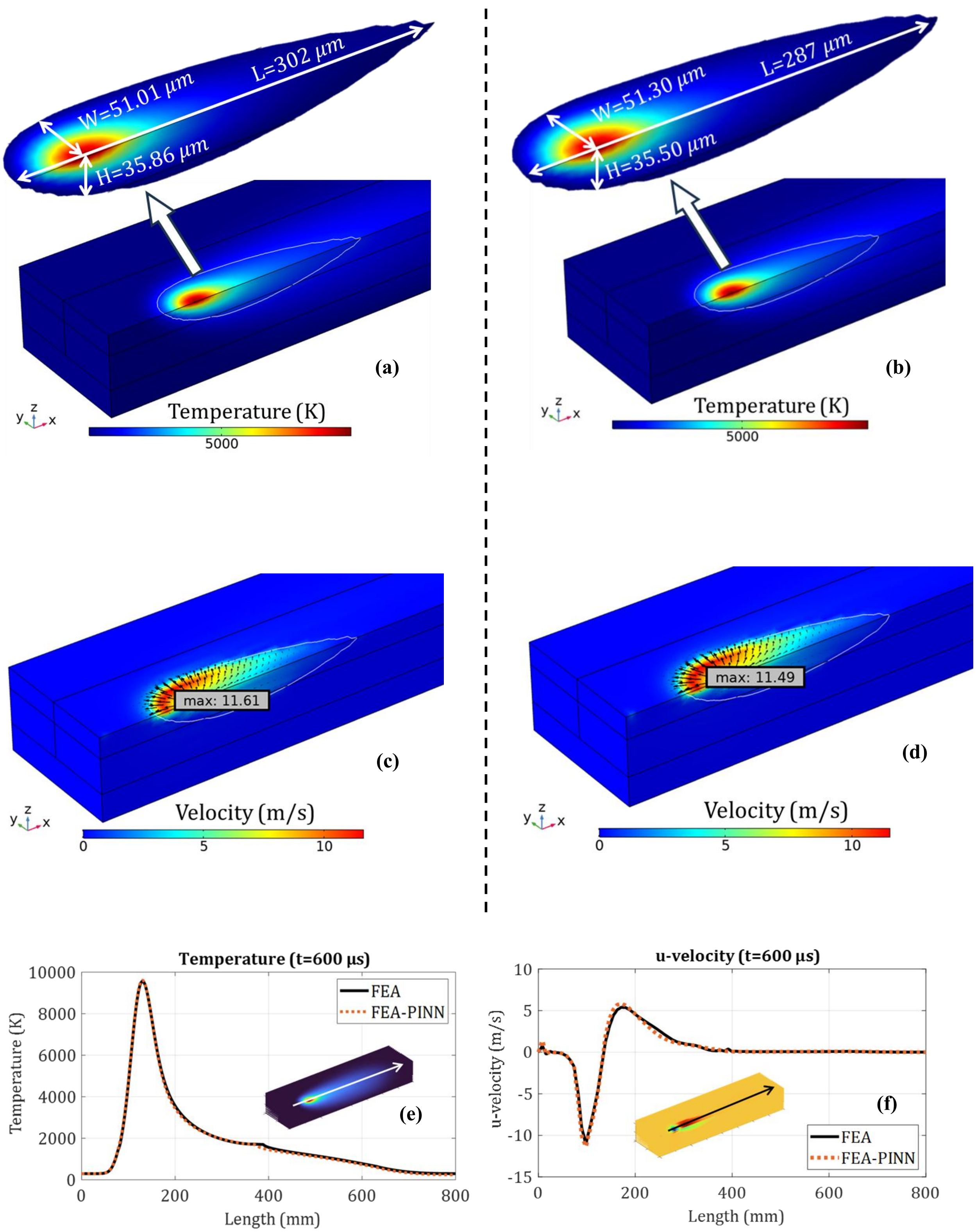

**Fig. 11:** Comparison between FEA and *FEA-PINN* model – melt pool temperature, shape, and resultant velocity.

Fig. 11 shows the zoomed view of the melt pool region to compare the *FEA-PINN* predicted temperature, shape, and velocity field with the FEA benchmark at t = 600 $\mu s$. The impact of the hybrid *FEA-PINN* approach is evident from Fig. 11, where the temperature and velocity predictions by the *FEA-PINN* model align closely with the benchmark FEA solution. The temperature distributions shown in Figs. 11(a) and 11(b) reveal that the *FEA-PINN* model almost accurately reproduces the maximum temperature of the melt pool, corresponding to an absolute error of 27.96 K (0.29%).

The extracted melt pool geometries in Figs. 11(a) and 11(b) show excellent agreement in all dimensions, with the predicted length (287 $\mu m$), width (51.30 $\mu m$), and height (35.50 $\mu m$) differing by less than 5% from the FEA results (302 $\mu m$, 51.01 $\mu m$, and 35.86 $\mu m$, respectively). The melt pool volumes also exhibit close correspondence, with the *FEA-PINN* model predicting 240,480 $\mu m^3$ compared to 244,660 $\mu m^3$ from the FEA simulation, showing a deviation of ~1.7%. This close match demonstrates the capability of the *FEA-PINN* model to accurately capture the melt pool morphology.

Similarly, the resultant velocity ($\sqrt{u^2 + v^2 + w^2}$) contours presented in Figs. 11(c) and 11(d) indicate that the predicted flow behavior within the molten pool is consistent with the FEA solution. The maximum velocity predicted by the *FEA-PINN* (11.49 m/s) deviates by only 1.03% from the FEA value (11.61 m/s), confirming the model's effectiveness in representing the Marangoni-driven convection.

Furthermore, in Figs. 11(e) and 11(f), the temperature and u-velocity profile obtained from *FEA-PINN* accurately match the FEA result, capturing both the peak temperature and the steep thermal gradient across the domain. This confirms the effectiveness of the *FEA-PINN* framework in restoring and maintaining accuracy over extended time durations.

Table 2 provides a comprehensive performance comparison of FEA, PINN, and *FEA-PINN* frameworks in terms of training data generation, training time, inference, correction loops, and overall accuracy. The FEA method provides high-fidelity results but demands substantial computational time (~11.75 hours), whereas the standalone PINN achieves rapid results in ~3.95 hours but suffers from accuracy degradation over time. The proposed *FEA-PINN* framework effectively balances these trade-offs, reducing the total simulation time to ~5.06 hours for the first time and ~3.64 hours for the parametric study by integrating PINN inference with periodic FEA-based corrections. As shown, three corrective loops are sufficient to refine the solution and restore physical consistency, demonstrating the framework's efficiency for large-scale LPBF simulation.

The authors acknowledge that the COMSOL simulations were executed on a CPU-based setup, as GPU acceleration is not yet fully supported for all physics interfaces within COMSOL Multiphysics at the time of this study. In contrast, the PINN model was trained and evaluated on an NVIDIA RTX A6000 GPU. Therefore, this comparison does not represent a strict one-to-one evaluation in terms of computational hardware. However, given the software's current limitations, the author utilized the best available computational resources to ensure a fair and realistic assessment of the proposed framework's efficiency. The authors also acknowledge that the LPBF simulation presented in this work is simplified, as it does not account for evaporation effects that typically lead to higher peak temperatures. Nevertheless, this study serves as a baseline framework for incorporating additional physical phenomena to achieve results that are more consistent with experimental observations. The primary objective of this work is to demonstrate a hybrid approach capable of accelerating complex Multiphysics simulations. The proposed strategy is particularly well-suited for scalable and computationally efficient modeling of LPBF thermal processes and other time-dependent problems. Moreover, it offers substantial potential for extension to more complex physics, such as evaporation, keyhole dynamics, and recoil pressure. The framework can also seamlessly integrate with commercial simulation tools to improve computational efficiency and enable high-fidelity, industrial-scale simulations.

**Table 2**. Comparison of FEA, PINN, and *FEA-PINN* model performance

| | **Task** | **FEA** | **PINN** | ***FEA-PINN*** |
|---|---|---|---|---|
| First set of process parameters | Training data generation | - | ~ 2.35h | ~ 2.35h |
| | Training + Inference time | - | ~ 1.6h | ~ 1.6h |
| | FEA-Correction 1 **(280 $\mu s$)** | - | - | ~ 0.25h |
| | PINN retraining + inference | - | - | ~ 0.18h |
| | FEA-Correction 3 **(440 $\mu s$)** | - | - | ~ 0.25h |
| | PINN retraining + inference | - | - | ~ 0.18h |
| | FEA-Correction 4 **(580 $\mu s$)** | - | - | ~ 0.25h |
| | Total time | ~ 11.75h | ~ 3.95h | ~ 5.06 |
| Other sets of process parameters (Transfer learning) | Training data generation | - | ~ 2.35h | ~ 2.35h |
| | Inference time | - | ~ 0.18h | ~ 0.18h |
| | FEA-Correction 1 **(280 $\mu s$)** | - | - | ~ 0.25h |
| | PINN retraining + inference | - | - | ~ 0.18h |
| | FEA-Correction 3 **(440 $\mu s$)** | - | - | ~ 0.25h |
| | PINN retraining + inference | - | - | ~ 0.18h |
| | FEA-Correction 4 **(580 $\mu s$)** | - | - | ~ 0.25h |
| | Total time | ~ 11.75h | ~ 2.53h | ~ 3.64h |

## 6. Conclusions

This work has developed a hybrid *FEA-regulated Physics-Informed Neural Network (FEA-PINN)* framework to address the challenges of error accumulation of conventional PINNs and high computational costs of physics-based simulations in LPBF. *FEA-PINN* combines the data efficiency and speed of PINNs with the physical reliability of periodic FEA corrections. The proposed PINN model incorporates temperature-dependent material properties, phase change of powder-liquid-solid through a novel dynamic material updating strategy, Marangoni and buoyancy-driven melt pool flow, and the framework leverages transfer learning to efficiently adapt to new process parameters. By dynamically integrating corrective FEA simulations during PINN inference, *FEA-PINN* provides a robust and practical approach to restore physical consistency and significantly reduce computational cost while maintaining accuracy comparable to traditional FEA methods for long-duration models. This hybrid approach holds great potential for future extension to more complex, multi-physics problems and integration into industrial applications, enhancing simulation and control of manufacturing processes. Key conclusions are summarized as follows.

- Conventional PINNs face major challenges in modeling complex, time-dependent LPBF processes due to microscale domain size, difficulty in normalizing the governing equations, phase-change dynamics, error accumulation, and limited labeled data requirements.
- A dynamic material updating strategy was implemented in the PINN model, using temperature history and a state variable to capture powder–liquid–solid transitions and represent temperature-dependent material properties accurately.
- As simulation horizon increases, the PINN model shows reduced accuracy, with lower peak temperatures and smaller melt pool size (t = 600 $\mu s$). The network smoothens sharp thermal gradients near the solid–liquid interface and exhibits noticeable errors in u-velocity prediction, highlighting its limitations in capturing steep temperature and flow variations.
- The *FEA-PINN* framework mitigates long-term modeling errors by incorporating periodic FEA corrections during inference, reducing residuals and restoring physical consistency to achieve accuracy comparable to full FEA simulations in single-track LPBF thermal fields.
- The *FEA-PINN* framework demonstrates excellent agreement with FEA in predicting LPBF thermal and flow fields, with a maximum temperature error of only 0.29% and velocity deviation of 1.03%.
- The predicted melt pool geometry shows close correspondence with FEA results, with dimensional differences under 5% and a volume deviation of only ~1.7%, confirming the model's accuracy in capturing melt pool morphology
- The *FEA-PINN* framework demonstrates significant computational efficiency, reducing total simulation time to ~5.06 hours compared to ~11.75 hours for the FEA solver in the initial run, and achieving rapid inference of ~3.64 hours using the pre-trained model for parametric studies, making it highly suitable for fast and iterative LPBF simulations.

In future work, the authors aim to extend the FEA-PINN framework by incorporating additional physics such as evaporation and keyhole dynamics to enhance its realism and predictive capability. Furthermore, the study will focus on evaluating computational efficiency and validating the accuracy of the pre-trained model against experimental data.

**Acknowledgments**

The authors would like to thank the financial support of the National Science Foundation under the grants CMMI-2152908 and ECCS-2328260.

**Data availability**

Data will be made available on request.